\def\year{2020}\relax
\documentclass[letterpaper]{article} 
\usepackage{aaai20}  
\usepackage{times}  
\usepackage{helvet} 
\usepackage{courier}  
\usepackage[hyphens]{url}  
\usepackage{graphicx} 
\usepackage{graphicx}  
\usepackage{latexsym}
\usepackage{bm}
\usepackage{amssymb}
\usepackage{amsmath}
\usepackage{multirow}

\newcommand{\citet}[1]{\citeauthor{#1} \shortcite{#1}} 
\newcommand{\citep}{\cite}

\title{Modeling Fluency and Faithfulness for Diverse Neural Machine Translation }
\author{Yang Feng\textsuperscript{\rm 1,2}~~~Wanying Xie\textsuperscript{\rm 1,3}~~~Shuhao Gu\textsuperscript{\rm 1,2}~~~Chenze Shao\textsuperscript{\rm 1,2}\\
\bf \Large Wen Zhang\textsuperscript{\rm 4}~~~Zhengxin Yang\textsuperscript{\rm 1,2}~~~Dong Yu\textsuperscript{\rm 3}\thanks{Corresponding author: Dong Yu. \newline \indent ~~ Our code can be got at https://github.com/ictnlp/DiverseNMT} \\ 
{\textsuperscript{\rm 1} {Key Laboratory of Intelligent Information Processing}} \\ Institute of Computing Technology, Chinese Academy of Sciences (ICT/CAS) \\
{ \textsuperscript{\rm 2} {University of Chinese Academy of Sciences, Beijing, China}} \\
{ \textsuperscript{\rm 3} Beijing Language and Culture University, China} \\
{ \textsuperscript{\rm 4} Smart Platform Product Department of Tencent Inc., China} \\
\{{fengyang, gushuhao19b, shaochenze18z, yangzhengxin17z}\}@ ict.ac.cn\\
xiewanying07@gmail.com,~~kevinwzhang@tencent.com,~~yudong@blcu.edu.cn
}

\begin{document}

\maketitle

\begin{abstract}
Neural machine translation models usually adopt the teacher forcing strategy for training which requires the predicted sequence matches ground truth word by word and forces the probability of each prediction to approach a 0-1 distribution. However, the strategy casts all the portion of the distribution to the ground truth word and ignores other words in the target vocabulary even when the ground truth word cannot dominate the distribution. To address the problem of teacher forcing, we propose a method to introduce an evaluation module to guide the distribution of the prediction. The evaluation module accesses each prediction from the perspectives of fluency and faithfulness to encourage the model to generate the word which has a fluent connection with its past and future translation and meanwhile tends to form a translation equivalent in meaning to the source. 
The experiments on multiple translation tasks show that our method can achieve significant improvements over strong baselines.
\end{abstract}

\section{Introduction}
Neural machine translation (NMT) \cite{kalchbrenner2013recurrent,sutskever2014sequence,bahdanau2014neural,gehring2017convolutional,vaswani2017attention,zhangwen2019bridging} has shown its superiority and drawn much attention recently. Most NMT models can fit in the attention-based encoder-decoder framework where the encoder projects the source sentence into representations in a common concept space and the decoder first retrieves related information from these representations and  then decodes it into target translation word by word. The training scenario is that each sentence is provided with a ground truth sequence and the teacher forcing strategy \cite{williams1989learning} is employed to force the generated translation to approach ground truth word by word via a cross-entropy loss. In this way,  the probability distribution of each prediction is expected to reach a 0-1 distribution with the ground truth word approaching to 1 and other words in the target vocabulary close to 0.

However, in practice the translation model cannot always generate the ground truth translation even with teacher forcing during training. One reason is that a source sentence can have multiple gold translations and hence in the training data there may be several different expressions for the same meaning segment. Another reason is that the model cannot fit to the training data perfectly due to noise and the expression ability of the model. But unfortunately teacher forcing ignores                                                                                                                                                                                                                                                                                                                                                                                                                                                                                                                                                                                                                                                                                                                                                                                                                                                                                                                                                                                                                              unreachable optimization situations where the ground truth word can only struggle to account for a small portion of the whole distribution, but still propagates the supervision only through the path from the ground truth word, having the major portion of the distribution excluded.

In this sense, it is beneficial to introduce an evaluation mechanism to guide the distribution when the prediction cannot converge to the ground truth word. If the prediction gives an alternative gold expression, the evaluation mechanism can give a proper evaluation to this word, otherwise, it can offer another distribution, rather than the 0-1 distribution, which can be seen as a lower optimization bound, to guide the training of the model. For evaluation, a translation is a good translation only when it can form a fluent sentence in the target and as well express the meaning of the source faithfully. Therefore, in the scenario of teacher forcing, each prediction should be evaluated from the perspectives of fluency and faithfulness. For the fluency,  the predicted word should be estimated whether to form a fluent sentence together with its past and future translation. For the faithfulness, the predicted word should be accessed whether to translate proper source information so that the whole source and target sequences have equivalent meanings. 

In this paper, we follow the above analysis to solve the problem of teacher forcing. Whenever generating a target word, we evaluate the fluency by connecting it with the self-generated past translation and the future ground truth translation and calculating the co-occurrence probability of the three parts. And we meanwhile estimate the faithfulness by first figuring out  the cross-attention over the source sentence with the past and future translation and then computing the probability of translating the attention into the target word. Furthermore, to trade off the fluency against the faithfulness, we integrate the two evaluation metrics together into a unified evaluation module and adjust their weights automatically. Finally the evaluation module is introduced to the NMT model to guide the distribution of predictions. The experiments on multiple translation tasks show that our method can outperform strong baselines significantly without any additional load for decoding. And the analysis experiments indicate that our method can have a better parameter fitting and produce more reasonable translation.



\section{Background} \label{sec-bg}

In this paper, we will introduce how to apply our method under the framework of {\em Transformer} \cite{vaswani2017attention} which has an encoder-decoder structure, so before diving into details, we will first introduce Transformer briefly. We denote the input sequence of symbols as $\bm{\mathrm{x}}=(x_1 ,..., x{_J})$,  the ground truth sequence as $\bm{\mathrm{y}}^*=(y^*_1 ,..., y^*_I)$ and the generated translation as $\bm{\mathrm{y}}=(y_1 ,..., y_I)$.

\subsection{The Encoder}

The encoder is composed of a stack of $N$ identical layers with each layer having two sublayers. The first sublayer is a multi-head attention unit used to compute the self-attention of the input, named {\em multi-head sublayer}, and the second is a fully connected feed-forward network, named {\em FNN sublayer}. Both of the sublayers are followed by a residual connection operation and a layer normalization operation. The multi-head attention unit  adopts dot-product attention which processes a set of queries ($\bm{\mathrm{Q}}$), keys ($\bm{\mathrm{K}}$) and values($\bm{\mathrm{V}}$) simultaneously, denoted as \
$\mathrm{MutiHead}(\bm{\mathrm{Q}}, \bm{\mathrm{K}}, \bm{\mathrm{V}}) $.

 For the $n$-th layer of the encoder, the multi-head sublayer can be formalized as
 \begin{equation}
\bm{\mathrm{Z}}^n={\mathrm{AddNorm}}({\mathrm{MutiHead}}(\bm{\mathrm{H}}^{n\mathcal{-}1}, \bm{\mathrm{H}}^{n\mathcal{-}1}, \bm{\mathrm{H}}^{n\mathcal{-}1})) \nonumber
\end{equation}
where $\bm{\mathrm{H}}^{n-1} \in  \mathbb{R}^{J \times d_\mathrm{model}}$ is the matrix of the packed output of the ${n\mathcal{-}1}$-th layer. A special case is that the queries, keys and values for the first layer of the encoder are all the matrix of the packed input embeddings $\bm{\mathrm{E}}_x=[{\mathrm{E}}_x[x_1]; ...; {\mathrm{E}}_x[x_J]]^T$ where ${\mathrm{E}}_x[x_j]$ 
is the sum of the embedding and position embedding of the source word ${\mathrm{x}}_j$.
 Then the FFN sublayer of the $n$-th layer is formalized as
 \begin{equation}
\bm{\mathrm{H}}^n={\mathrm{AddNorm}}({\mathrm{FFN}}(\bm{\mathrm{Z}}^n)) .
\end{equation}

Then the output of the $N$-th layer is taken as source hidden states and we denote its packed matrix as $\bm{\mathrm{H}}$.
 
 \subsection{The Decoder}

The decoder is also composed of a stack of $N$ identical layers. For each layer, besides the multi-head sublayer and the FFN sublayer,  a third sublayer is inserted, called {\em cross-attention sublayer}. The cross-attention sublayer performs multi-head attention over source hidden states with the output of the multi-head sublayer in the same layer as query. Residual connection and layer normalization are also applied after each sublayer. In addition, as we do not know the future translation, a mask matrix is applied to prevent the subsequent target words from being involved. 

Formally, for the $n$-th layer of the decoder,  the multi-head sublayer is denoted as 
\begin{equation}
\bm{\mathrm{A}}^n={\mathrm{AddNorm}}({\mathrm{MutiHead}}(\bm{\mathrm{S}}^{n\mathcal{-}1}, \bm{\mathrm{S}}^{n\mathcal{-}1}, \bm{\mathrm{S}}^{n\mathcal{-}1}) \nonumber
\end{equation}
where $\bm{\mathrm{S}}^{n-1} \in  \mathbb{R}^{I \times d_\mathrm{model}}$ is the output of the $n\mathcal{-}1$-th layer and specifically the queries, keys and values for the first layer are all the target embedding matrix  $\bm{\mathrm{E}}_{y}$.
The cross-attention sublayer is written as 
\begin{equation}
\bm{\mathrm{C}}^n={\mathrm{AddNorm}}({\mathrm{MutiHead}}(\bm{\mathrm{A}}^{n}, \bm{\mathrm{H}}, \bm{\mathrm{H}}) 
\end{equation}
and the FFN sublayer is formalized as 
\begin{equation}
\bm{\mathrm{S}}^n={\mathrm{AddNorm}}({\mathrm{FFN}}(\bm{\mathrm{C}}^{n}) ) \ .
\end{equation}
After these operations, the final output of the $N$-th layer gives the target hidden states, denoted as  $\bm{\mathrm{S}}=[\bm{\mathrm{s}}_1; ...; \bm{\mathrm{s}}_I]^T$, where $\bm{\mathrm{s}}_i$ is the hidden state of $y_i$. 

By performing a linear transformation and a softmax operation to the target hidden states, we can get the translation probability as
\begin{equation}
p(y_i | \bm{\mathrm{y}}_{<i}, \bm{\mathrm{x}}) \propto \exp{(\bm{\mathrm{s}}_i \bm{\mathrm{W}}_o)} \label{eq-outprojection}
\end{equation}
where  $\bm{\mathrm{W}}_o \in  \mathbb{R}^{d_\mathrm{model} \times |V_t|}$ and $|V_t|$ is the size of the target vocabulary.

Transformer is trained by minimizing a cross-entropy loss which maximizes the probability of the ground truth sequence:
\begin{equation}
\mathcal{L}=-\sum_{i=1}^{I} \log p(y^*_i | \bm{\mathrm{y}}_{<i}, \bm{\mathrm{x}})
\end{equation}

\begin{figure}[!t]
    \centering
   \includegraphics[width=0.6\columnwidth]{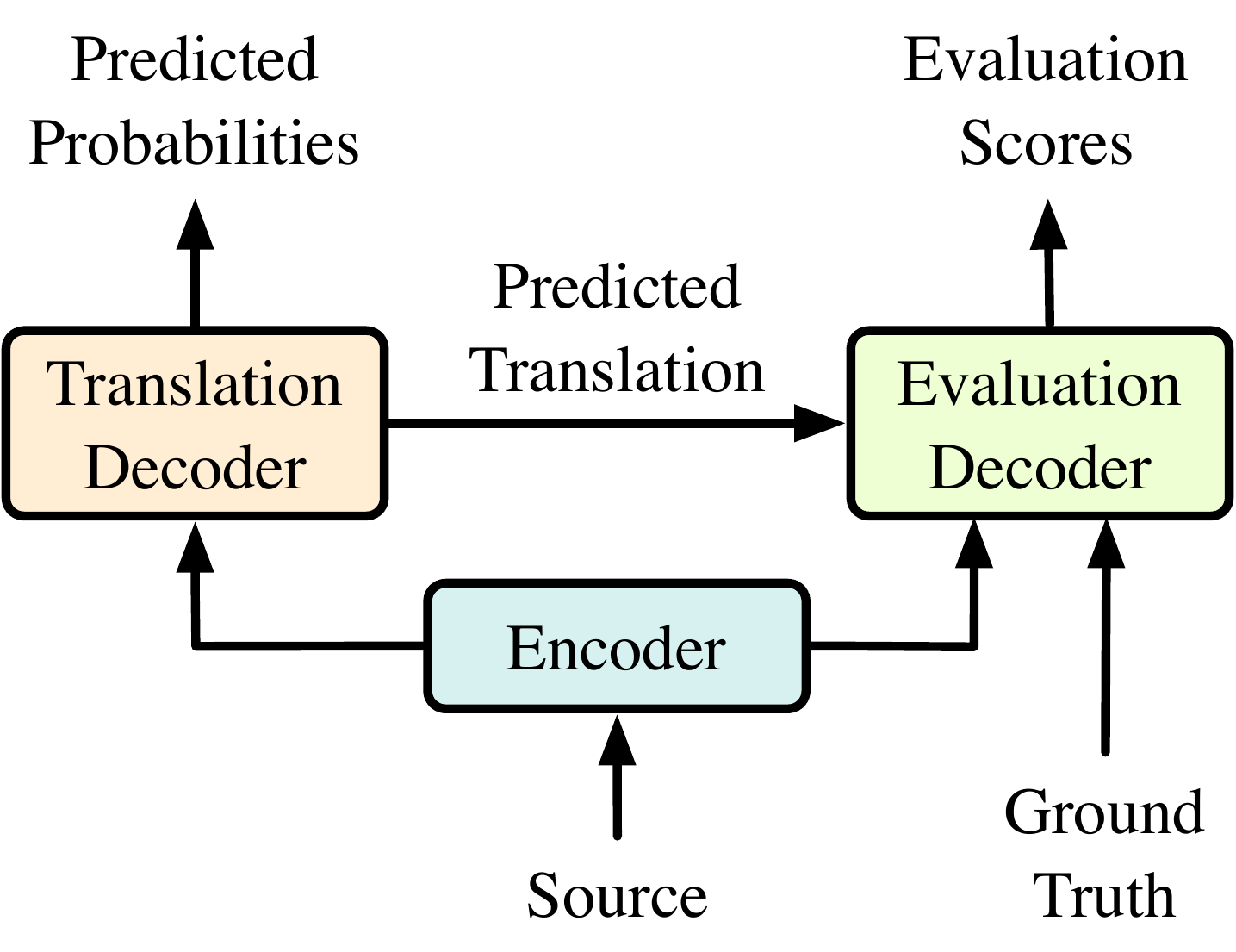}
    \caption{The architecture of the proposed method.}
    \label{fig-model}
\end{figure}

\section{Model}

Our work aims to introduce an evaluation module into the NMT model to provide a more reachable distribution to fit, as a complement to the 0-1 distribution adopted by the cross-entropy loss, so our model consists of two modules. One is {\em the translation module}, composed of the encoder and the translation decoder, which is used to generate candidate translations in the same way as Transformer. The other is {\em the evaluation module}, containing the encoder and the evaluation decoder, which is used to evaluate the translation produced by the translation module word by word. The whole architecture is shown in Figure \ref{fig-model}. Please note that the encoder is shared by the two decoders. Then the distribution drawn by the evaluation module is used in an additional loss to guide the distribution drawn by the translation module. Then at test our model leaves out the evaluation module and performs inference only with the translation module.

\begin{figure*}[!ht]
    \centering
   \includegraphics[width=1.35\columnwidth]{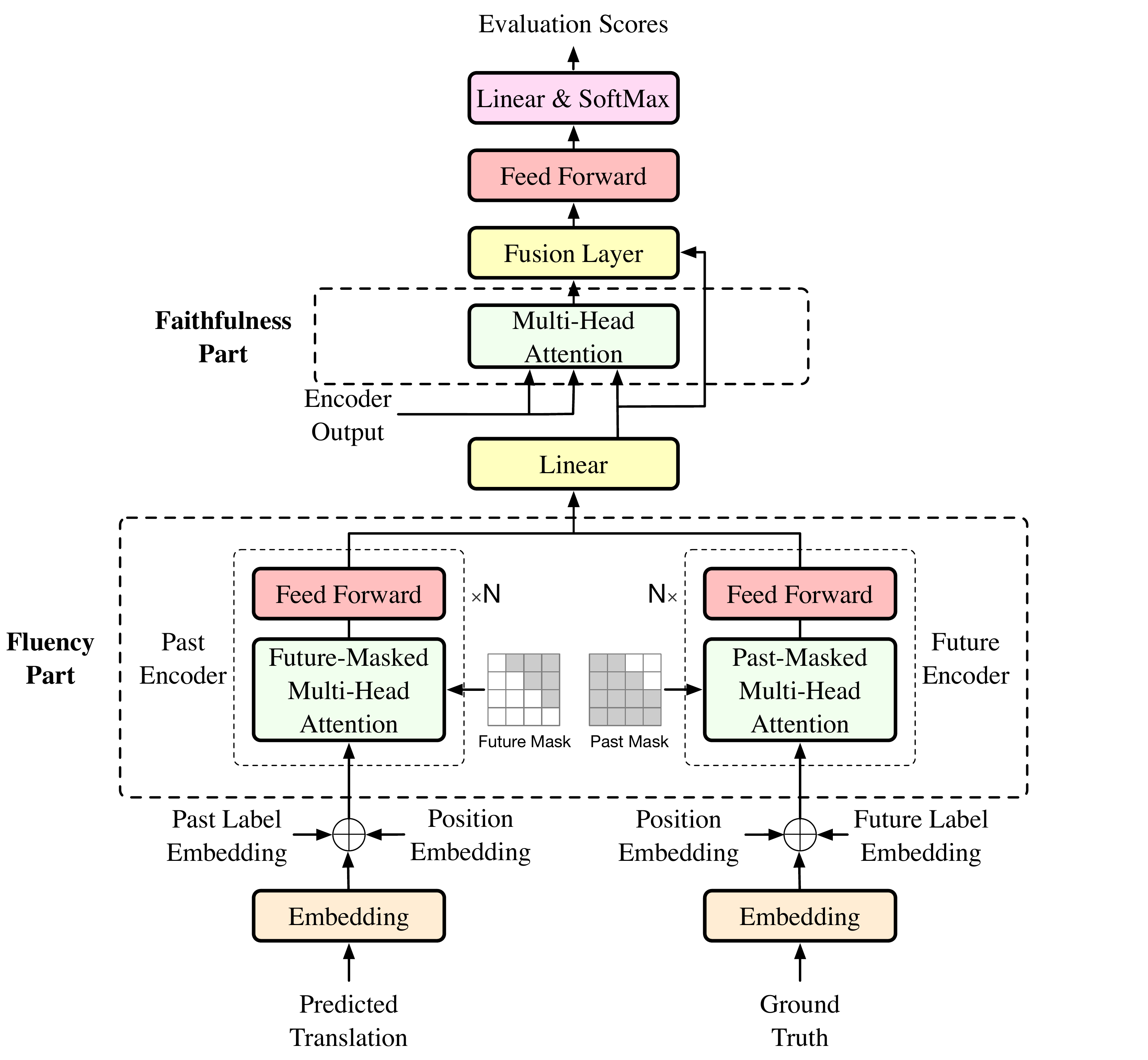}
    \caption{The architecture of the evaluation module.}
    \label{fig-evaluation}
\end{figure*}

\subsection{The Evaluation Module}

In our method, whenever the translation module generates a target word, the evaluation module will assess it from the perspectives of fluency and faithfulness. The two metrics are integrated together to come out with a final score so as to trade off against each other. Thus the evaluation module can be logically divided into three parts as the fluency part, the faithfulness part and the fusion layer. The architecture is shown in Figure \ref{fig-evaluation}. 


{\bf The Fluency Part}

To ensure the fluency of the whole translation, each generated word should be a good conjunction with its past and future translation. At the time step $i$, we use as {\em the past translation} the translation predicted by the translation module, denoted as $y_1, ..., y_{i-1}$. Under the training framework of teacher forcing, in the following steps the model will be taught to generate the rest of the ground truth sequence $y^*_{i+1}, ..., y^*_I$, so we use this as {\em the future translation}. Please note that the past translation is predicted during training which means the context rolled in the decoder is ground truth words, not self-generated words.

Then to get the representation for the past translation and the future translation, we employ two encoders, called {\em past encoder} and {\em future encoder}. Then the fluency of the word $y_i$ can be assessed conditioned on the representations of the past and future information, just like the evaluation of language models.

The past encoder consists of a stack of $N$ identical layers with each layer having two sublayers of a multi-head sublayer and an FFN sublayer, just like the encoder of Transformer.  
In order to support parallel training, we first collect the whole sequence generated by the translation module with ground truth words as context, denoted as $y_1, ..., y_I$, and then feed the whole sequence to the past encoder. Then to pick the past translation at each time step $i$, we mask out the current and future translation $y_i, ..., y_I$ in the multi-head sublayer,  corresponding to {\em future-masked multi-head attention} in Figure \ref{fig-evaluation}. 

The future encoder has the same structure except that its input is the whole ground truth sequence and at the time step $i$, the masked part is the past and current ground truth words $y^*_1, ..., y^*_i$. Its multi-layer sublayer corresponds to  {\em past-masked multi-head attention} in Figure \ref{fig-evaluation}.

In addition, besides the word embedding and the position embedding, we introduce a new label embedding with two labels to indicate whether a word comes from a past translation or a future translation. Then the final embeddings fed to the two encoders are the sum of the three embeddings.

Assume the hidden state matrices outputted by the past and future encoders are $\bm{\mathrm{A}}_p$ and $\bm{\mathrm{A}}_f$, respectively, then the two outputs are fused together to produce the condition $\bm{\mathrm{A}}_e$ for fluency estimation as
\begin{equation}
\bm{\mathrm{A}}_e = \bm{\mathrm{W}}_p \bm{\mathrm{A}}_p + \bm{\mathrm{W}}_f \bm{\mathrm{A}}_f
\end{equation}
where $\bm{\mathrm{W}}_p$ and $\bm{\mathrm{W}}_f$ are linear transformations.

{\bf The Faithfulness Part} 

In addition to fluency, the generated word should also reflect a proper amount of source meaning so that the whole translation can express the source sentence faithfully. Here faithfulness means being adequate and accurate in meaning. We model the evaluation for faithfulness as a translation task and estimate the translation probability (faithfulness) word by word. The scenario is to first retrieve the related source information and then to assess the probability of translating the retrieved information to the given target word.

To figure out the corresponding source information, we perform the cross-attention over the source hidden states $\bm{\mathrm{H}}$ generated by the shared encoder, using as the query the fusion of the past and future information $\bm{\mathrm{A}}_e$. 
This can be formalized as
\begin{equation}
\bm{\mathrm{C}}_e={\mathrm{AddNorm}}({\mathrm{MutiHead}}(\bm{\mathrm{A}}_{e}, \bm{\mathrm{H}}, \bm{\mathrm{H}}) )
\end{equation} 
where $\bm{\mathrm{C}}_e$ is the generated cross-attention. Then the translation probability is calculated in the same way as Transformer does.


{\bf The Fusion Layer} 

As the generated translation is assessed with two metrics of fluency and faithfulness, the two metrics should be traded off against each other. For the two metrics both aim to get a conditional probability for the current generated word, we fuse the conditions of them, $\bm{\mathrm{A}}_e$ and $\bm{\mathrm{C}}_e$, to get a new combined condition which is followed by an FFN layer. This process can be written as
\begin{gather}
\bm{\mathrm{B}}_e = \bm{\mathrm{W}}_a \bm{\mathrm{A}}_e + \bm{\mathrm{W}}_c \bm{\mathrm{C}}_e \\
\bm{\mathrm{S}}_e = {\mathrm{AddNorm}}({\mathrm{FFN}}(\bm{\mathrm{B}}_{e}) )
\end{gather}
where $\bm{\mathrm{S}}_e=[\bm{\mathrm{s}}_{e1}; ...; \bm{\mathrm{s}}_{eI}]^T$ and $\bm{\mathrm{s}}_{ei}$ is the hidden state for the word $y_i$. Then with a linear transformation $\bm{\mathrm{W}}_e$ and a softmax operation, we can get the final evaluation score for $y_i$ as
\begin{equation}
p_e(y_i | \bm{\mathrm{y}}^*_{>i}, \bm{\mathrm{y}}_{<i}, \bm{\mathrm{x}}) \propto \exp{(\bm{\mathrm{s}}_{ei} \bm{\mathrm{W}}_e)} \label{eq-outprojection}
\end{equation}

\subsection{Training}

During training, our method not only jointly optimizes the translation module and the evaluation module but employs an additional loss to guide the behavior of the translation module.
Specifically, for the translation module, a cross-entropy loss is employed as
\begin{equation}
\mathcal{L}_t=-\sum_{k=1}^{K} \sum_{i=1}^{I} \log p(y^*_i | \bm{\mathrm{y}}_{<i}, \bm{\mathrm{x}}) \ .
\end{equation}
The evaluation module is also optimized via a cross-entropy loss as 
\begin{equation}
\mathcal{L}_e=-\sum_{k=1}^{K} \sum_{i=1}^{I} \log p_e(y^*_i | \bm{\mathrm{y}}^*_{>i}, \bm{\mathrm{y}}_{<i}, \bm{\mathrm{x}}) \ .
\end{equation}

For the correlation of the two modules,  a common practice is to employ the Kullback-Leibler (KL) divergence  as the loss to make sure the distributions drawn by the two modules get close to each other. However, it is not optimal to bind the two distributions over the whole target vocabulary as this will hinder the model to search for a better minimum. Instead, we only pay attention to the word generated by the translation module and use the evaluation module to guide the probability given by the translation module. This loss is given as
\begin{equation}
\mathcal{L}_\mathrm{c}=\sum_{k=1}^{K} \sum_{i=1}^{I} p_e(y_i | \bm{\mathrm{y}}^*_{>i}, \bm{\mathrm{y}}_{<i},\bm{\mathrm{x}}) \log p(y_i | \bm{\mathrm{y}}_{<i}, \bm{\mathrm{x}}) \ .
\end{equation}
With this loss, if the generated word happens to be the ground truth word, then the distribution drawn by the translation module will be sharper at the ground truth word, otherwise, the translation module tends to reinforce the translation with higher confidence given by the evaluation module. In the experiment section, we will verify that $\mathcal{L}_\mathrm{c}$ can bring about better performance than the KL divergence.

The final loss is  
\begin{equation}
\mathcal{L}= \mathcal{L}_t +\mathcal{L}_e + \mathcal{L}_\mathrm{c} \label{eq-loss}
\end{equation}

In the training, we first pretrain the translation and evaluation modules together with the loss $\mathcal{L}_{\mathrm{pretrain}}= \mathcal{L}_t +\mathcal{L}_e$. Near convergency, we introduce $\mathcal{L}_\mathrm{c}$ and fine tune the model with the loss in Equation \ref{eq-loss}.

\section{Related Work}

Our work introduces an evaluation module to give an assessment to the predicted word so that it can always have gradient back propagated through. Some researchers also took effort in this direction. \citet{shao2018greedy} employed a greedy search strategy to generate translation, then used the accuracy of probabilistic n-grams as the training loss. This method calculates the probabilities of n-grams by counting probabilistic occurrences through the entire vocabulary, hence it has all the words involved in the optimization with the sequence-level n-gram loss.
\citet{yang2019sentence} came out with a sentence-level agreement loss to directly model the difference between the representation of the source and target sentences, so that the source representation could be enhanced. \citet{wieting2019beyond} used a semantic similarity loss as a reward to measure the similarity between the embedding of the generated translation and the reference. \citet{elbayad2018token} extended the training loss with a token-level and sequence-level smoothing loss which smooths the target distribution over similar sentences.

In our method, the self-generated translation other than the ground truth can be involved during gradient back propagation. Some other work also tries to have more translation to take part in parameter update. These work usually adopts the REINFORCE algorithm \cite{williams1992simple} and samples translation to optimize according the probability distribution. 
This series of work \cite{wu2018study,yang2018improving,geng2018adaptive,kreutzer2018reliability} samples a translation from all the possible translation and performs gradient descent through the translation with a sequence-level reward under the framework of reinforcement learning .

Our method provides evaluation metrics for translation different from ground truth, allowing for diverse translation. There are also some other work which encourages diverse translation. \citet{ma2018bag} presented a bag-of-word loss for the model to generate translation with words in ground truth but in flexible word order.
 \citet{he2018sequence} developed a sequence-to-sequence mixture model to adopt a committee of translation models. Each translation model selects its own training set via optimization of marginal log-likelihood, leading to a soft clustering of the training data. By this mean, the method can improve the diversity and quality. \citet{shu2019generating} attempted to generate diverse translation by first extract sentence codes with or without syntax information and then sampling translation based on the sentence codes.

Our method evaluates from the perspectives of fluency and faithfulness and from this point of view, the line of work which first generate future translation also work for better fluency. \citet{hassan2018achieving} and \citet{zhang2018asynchronous} proposed a two-pass decoding algorithm which first generated a translation draft then refined while \citet{zhang2019synchronous} proposed to decode forwards and backwards simultaneously.
The main difference from our work is that they require decoding decoding bi-directionally while our method can decode once just like Transformer. Although \citet{serdyuk2018twin} also employ bidirectional decoding during training, they instead use the backward decoder to assistant the forward decoder, so that the forward decoder can generate hidden states close to the backward decoder. In this way, the forward decoder can produce similar translation to the backward decoder in the rest part and hence the backward decoder can be abandoned at test.

\section{Experiments}

In the experiment section, we will first report the comparison results with other strong baselines, then analyze the importance of all the factors in the model, then show the upper bound of the performance of the evaluation module, and next verify whether our method can achieve better optimization. Finally, we will indicate whether our method can generate translation of better fluency and faithfulness.


\begin{table*}[ht!]
\centering
\resizebox{2.1\columnwidth}!{
\begin{tabular}{l | l l l l l l l   |   l l   |   l l}
\hline
& \multicolumn{7}{|c|}{ \bf{CN$\rightarrow$EN} } & \multicolumn{2}{|c|}{\bf{EN$\rightarrow$DE}} & \multicolumn{2}{|c}{\bf{EN$\rightarrow$RO}} \\ 
& \bf \textbf{MT03} & \bf \textsc{MT04} & \bf \textsc{MT05} &  \bf \textsc{MT06}  & \bf \textsc{MT08} & \bf \textsc{AVE}  & \bf $\small{\Delta}$ & \bf \textsc{WMT14} & \bf $\Delta$  & \bf \textsc{WMT16} & \bf $\Delta$ \\
\hline
\hline
\bf \textsc{Transformer}  & 44.74 & 46.27 & 44.16 & 43.29 & 34.72 & 42.63 &  & 27.21 &   & 32.85 &   \\
\bf \textsc{\quad +RL}  & 44.50 & 45.96 & 44.26 & 43.92 & 35.55 & 42.83 &  {\em \small{+0.20}}  & 27.25 &  {\em \small{+0.04}} & 33.00 &  {\em \small{+0.15}} \\
\bf \textsc{\quad +BOW}  & 44.59 & 46.40 & 45.03 & 43.91 & 35.31 & 43.04 &  {\em \small{+0.41}}  & 27.35 &  {\em \small{+0.14}} & 32.95 &  {\em \small{+0.10}} \\
\bf {Our Method}-KL & 45.17 &	46.86** & 45.01** & 44.51* & 36.03* & 43.51 & {\em \small{+0.88}} & \bf{27.55} & \bf{\em \small{+0.34}} & 33.44 & {\em \small{+0.59}} \\
\bf {Our Method} & \bf{46.20*} & \bf{47.39*} & \bf{46.22*} & \bf{45.63*} & \bf{36.78*} & \bf{44.44} &  \bf{\em \small{+1.81}} & 27.35 &  {\em \small{+0.14}} & \bf{34.00*} &  \bf{\em \small{+1.15}} \\
\hline
\end{tabular}
}
\caption{BLEU scores on three translation tasks. *  and ** mean the improvements over  \textsc{Transformer} is statistically significant \cite{collins2005clause} ($\rho < 0.01$ and $\rho < 0.05$, respectively).} \label{tab-total}
\end{table*}

\begin{table*}[t!]
\centering
\resizebox{1.25\columnwidth}!{
\begin{tabular}{l | c | c | c | c | c | c}
\hline
& \bf \textbf{\small MT03} & \bf \textsc{\small MT04} & \bf \textsc{\small MT05} &  \bf \textsc{\small MT06}  & \bf \textsc{\small MT08} & \bf \textsc{\small AVE} \\
\hline
\hline
\bf{Full} & 46.20 & 47.39 & 46.22 & 45.63 & 36.78 & 44.44 \\
\hline
\bf {\; -Faithfulness}  & 44.95 & 46.85 & 45.15 & 44.45	 & 36.18 & 43.51 \\
\bf {\; -$\mathcal{L}_\mathrm{c}$} & 44.93 & 45.77 & 44.61 & 44.76 & 35.87 & 43.18 \\
\bf {\;  \; -Evaluation}  & 44.74 & 46.27 & 44.16 & 43.29 & 34.72 & 42.63 \\
\hline
\end{tabular} 
}
    \caption{Ablation study on the CN$\rightarrow$EN translation task. {Full}: our full model. {-Faithfulness}: deleting cross-attention from the faithfulness part.  {-$\mathcal{L}_\mathrm{c}$}: erasing $\mathcal{L}_\mathrm{c}$ in Equation \ref{eq-loss}. {\; -Evaluation}: removing the evaluation decoder (degrading to Transformer).} \label{tab-ablation}
\end{table*}

\begin{table*}[t!]
\centering
\resizebox{1.25\columnwidth}!{
\begin{tabular}{l | c  c  c  c  c  c | c}
\hline
& \bf \textbf{\small MT03} & \bf \textsc{\small MT04} & \bf \textsc{\small MT05} &  \bf \textsc{\small MT06}  & \bf \textsc{\small MT08} & \bf \textsc{\small AVE} & \bf{\small EN$\rightarrow$DE}\\
\hline
\hline
\bf Tran  & 31.31 & 23.87 & 29.02 & 30.61 & 22.26 & 27.41 & 31.76\\
\hline
\bf Eval  & 45.86 & 39.34 & 44.18 & 45.21 & 36.02 & 42.12 & 34.56 \\
\hline
\end{tabular} 
}
\caption{Performance comparison of the translation and evaluation modules on the CN$\rightarrow$EN and EN$\rightarrow$DE translation tasks. Ground truth is fed to both the modules and only one reference was used for the CN$\rightarrow$EN translation.} \label{tab-eval}
\end{table*}

\subsection{Data Preparation}

We conducted experiments on the following three data sets.

{\bf CN$\rightarrow$EN} The training data consists of 1.25M sentence pairs from LDC corpora which has 27.9M Chinese words and 34.5M English words respectively \footnote{The corpora include LDC2002E18, LDC2003E07, LDC2003E14, Hansards portion of LDC2004T07, LDC2004T08 and LDC2005T06.}. The data set MT02 is used as validation and MT03, MT04, MT05, MT06, MT08 are used for test. We tokenized and lowercased English sentences using the Moses scripts\footnote{http://www.statmt.org/moses/}, and segmented the Chinese sentences with the Stanford Segmentor\footnote{https://nlp.stanford.edu/}. The two sides were further segmented into subword units using Byte-Pair Encoding(BPE)~\cite{sennrich2016neural} with $30$K merge operations.

{\bf EN$\rightarrow$DE} The training data is from WMT2014 which consists about 4.5M sentences pairs with 118M English words and 111M German words. We chose the news test-2013 for validation and news-test 2014 for test.  BPE was also employed with $32$K merge operations.

{\bf EN$\rightarrow$RO} We used the preprocessed version of WMT16 English-Romanian dataset released by \citet{lee2018deterministic} which includes 0.6M sentence pairs. We use news-dev 2016 for validation and news-test 2016 for test. The two languages share the same vocabulary generated with $40$K merge operations of BPE.

\subsection{Systems}

We conducted our experiments based on self-attention-based encoder-decoder frame.



{\bf \textsc{Transformer}}  An open-source toolkit called {\em Fairseq-py} released by Facebook \cite{edunov2017fairseq} which was implemented strictly referring to \citet{vaswani2017attention}. 

{\bf +RL} Transformer trained under the reinforcement learning framework with the BLEU as the rewards, specifically the REINFORCE algorithm \cite{williams1992simple}. The implementation details for the RL part is the same as \citet{yang2018improving}.

{\bf +BOW} Our implementation of \citet{ma2018bag} on the basis of Transformer.

{\bf Our Method-KL} Implemented based on Fairseq-py. For the evaluation module, the fluency part is composed of a stack of $N$ = 6 layers. And the final loss for this system is 
\begin{equation}
\mathcal{L}= \mathcal{L}_t +\mathcal{L}_e + \mathcal{L}_\mathrm{KL} \nonumber
\end{equation}
where 
\begin{equation}
\mathcal{L}_\mathrm{KL}=\sum_{k=1}^{K}  \sum_{\bm{\mathrm{y}}_{i} \neq \bm{\mathrm{y}}^*_{i}} \mathrm{D_{KL}} (p_e(y_i | \bm{\mathrm{y}}^*_{>i}, \bm{\mathrm{y}}_{<i}, \bm{\mathrm{x}}) || p(y_i | \bm{\mathrm{y}}_{<i}, \bm{\mathrm{x}}) ) \ . \nonumber
\end{equation}
The KL loss forces the distributions drawn by the translation module and the evaluation module to approach to each other when the word generated by the translation module is different from the ground truth word.

{\bf Our Method} Implemented the same as the system {\em Our Method-KL} except that its final loss is $\mathcal{L}= \mathcal{L}_t +\mathcal{L}_e + \mathcal{L}_\mathrm{c}$ as shown in Equation \ref{eq-loss}.

All the Transformer-based systems have the same configuration as the base model described in \citet{vaswani2017attention}.

The translation quality was evaluated using the {\em multi-bleu.pl} scipt  \cite{papinenibleu} based on case-insensitive $n$-gram matching with $n$ up to $4$.

\subsection{Performance}


We compare with methods using different losses, such as \citet{yang2018improving} (named \textsc{+RL}) which gives a reward to all the possible translation and perform policy gradient propagation via REINFORCE algorithm, and  \citet{ma2018bag} (named \textsc{+BOW}) which encourage diverse translation via a bag-of-words loss. We also compared the two different additional losses previously mentioned $\mathcal{L}_\mathrm{KL}$ and $\mathcal{L}_\mathrm{c}$.

The results are shown in Table \ref{tab-total}. We can see that 
the improvement brought by \textsc{+RL} and \textsc{+BOW} methods are not great. For the \textsc{+RL} method, as we all know, the training is not stable and  thus it is difficult to converge to good optima. For the \textsc{+BOW}, the reason may be it does not model the word order in the training loss and hence has a loose supervision to the fluency. On the CN$\rightarrow$EN and EN$\rightarrow$RO translation tasks, the loss $\mathcal{L}_\mathrm{c}$ is much more effective than $\mathcal{L}_\mathrm{KL}$ while on the EN$\rightarrow$DE translation the result is reversed.
Another finding is that all the methods cannot achieve big improvements on the EN$\rightarrow$DE translation. The reason may be the training data is big enough and the two languages EN and DE are closed to each other. As a result, Transformer has already got a good enough optimization and it is difficult for further improvements. Our method with $\mathcal{L}_\mathrm{c}$ as an additional loss can outperform all the baselines significantly on the CN$\rightarrow$EN and EN$\rightarrow$RO translation tasks. Therefore, we can conclude that the evaluation module can help improve translation performance and the loss $\mathcal{L}_\mathrm{c}$ is more reasonable.


\subsection{Ablation Study}

Our method introduces three new factors into Transformer via the evaluation module, including the cross attention for faithfulness,  past and future encoders for fluency and the additional loss $\mathcal{L}_\mathrm{c}$. Here we conducted experiments to check their influence to our method by leaving them out one by one. The results are given in Table \ref{tab-ablation}.

We first left out the cross-attention unit which means we discard faithfulness and only consider fluency in our method, and find that the translation performance decreases greatly. This is in line with our conjecture that a good translation should be ensured to be faithful with the source meaning. 
We also tried to only abandon the loss $\mathcal{L}_\mathrm{c}$ in Equation \ref{eq-loss} and now the evaluation module can only participate in the optimization of the shared encoder (referring to Figure \ref{fig-model}). We can find the performance declines most of all. This is not difficult to understand.
When the predicted word is different from the ground truth word, the probability of the ground truth word only accounts for a small portion of the distribution and the cross-entropy loss distributing the rest great portion to 0 is not reasonable. Next we excluded the whole evaluation decoder, the performance further declines. This indicates that even we do not employ the distribution generated by the evaluation module to directly guide the behaviors of the translation module, the evaluation decoder can help seek better parameters for the share encoder.

\subsection{The Performance of the Evaluation Module}

As we use the evaluation module to guide the probability distribution of the translation model, to make sure the reasonability of this mechanism, the evaluation module should have an obvious superiority in the performance over the translation module. We conducted experiments to check this. For the evaluation module requires the participation of ground truth, we fed the reference of the test set to the evaluation module, so that the evaluation module can assess the fluency with self-generated translation as past translation and ground truth as future translation. For the CN$\rightarrow$EN translation, we only selected one reference to feed and calculated BLEU scores based on this single reference. In order to compare fairly, we also fed reference to the translation module to  use it as context.

According to the results in Table \ref{tab-eval}, we can find that the evaluation module indeed has a great margin in performance over the translation module consistently on all the test sets. The margin is bigger on the CN$\rightarrow$EN translation than on the EN$\rightarrow$DE translation which also gives an explanation to why the improvement on the CN$\rightarrow$EN is much bigger. 
Then we can conclude that it is reasonable to leverage the evaluation module to help optimize the translation module.

\subsection{The Reasonability of the Loss}

Our method uses the evaluation module to introduce a new distribution, then it adds $\mathcal{L}_\mathrm{c}$ to the loss, aiming to achieve a better optimization with the help of the new distribution. We design experiments to check this in two aspects. First is whether the loss in Equation \ref{eq-loss} is reasonable, which means lower loss can lead to better translation (greater BLEU scores) on the training data. Second is whether the loss can result in better optimization, that is higher BLEU scores on the valid data.

The experiment details are as follows. We first pretrained our method with the loss $\mathcal{L}_{\mathrm{pretrain}}= \mathcal{L}_t +\mathcal{L}_e$ for the first 10 epochs and then added $\mathcal{L}_\mathrm{c}$ to the loss afterwards. Then we sampled 1000 sentences from the training set and tested the BLEU score on sampled training set with ground truth words as context. In this way, we can see whether the translation module behaves as we want. We also tested BLEU scores on the valid set with self-generated translation as context. 

\begin{figure}[!t]
    \centering
   \includegraphics[width=0.9\columnwidth]{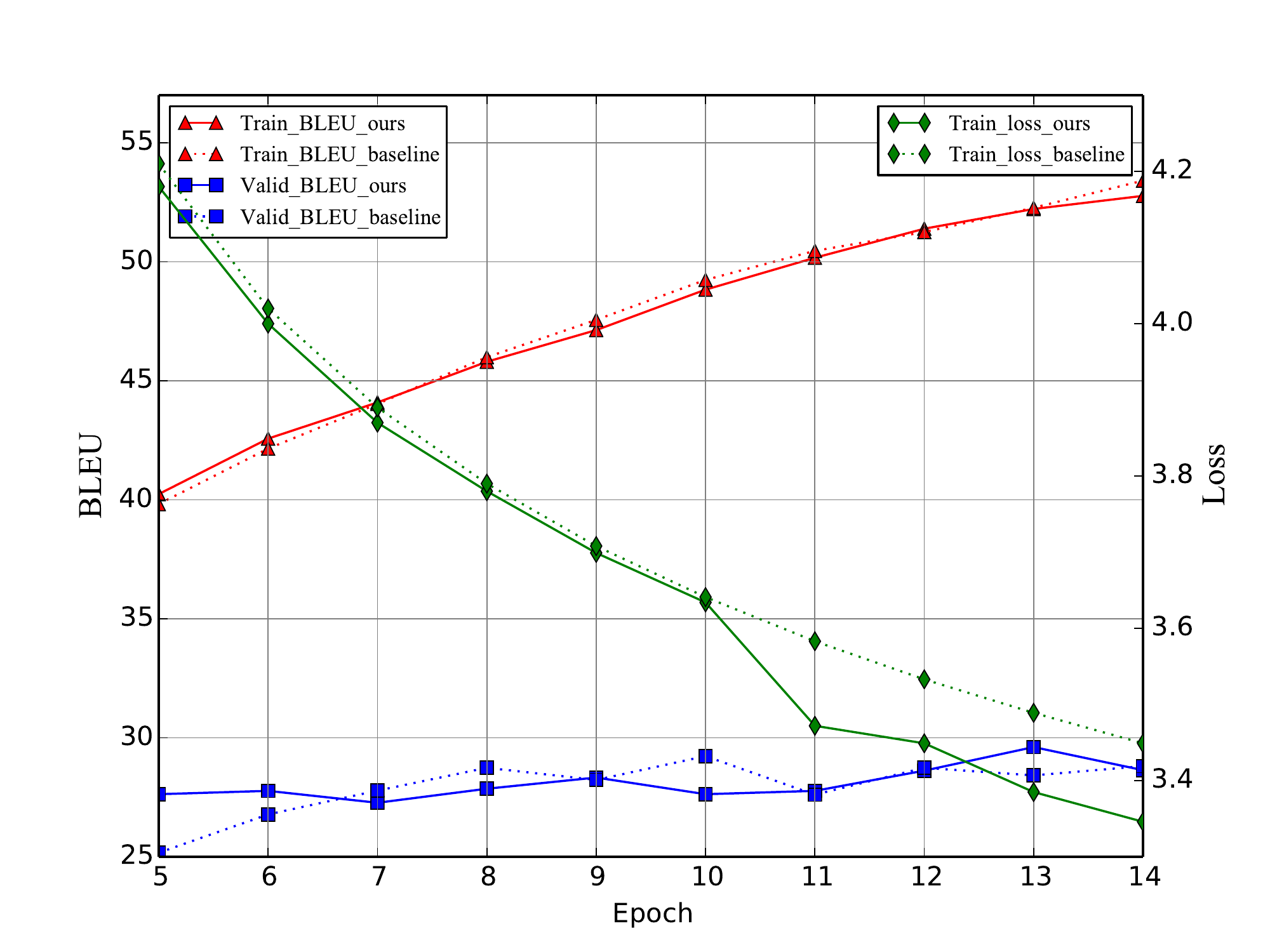}
    \caption{Training losses and BLEU scores on the CN$\rightarrow$EN training and valid sets.}
    \label{fig-converg}
\end{figure}

We put training losses and BLEU scores all in the Figure \ref{fig-converg} with two scalars for y-axis. From the results, we can see that the training loss of our method decreases gradually before $\mathcal{L}_\mathrm{c}$ is added (epoch 10) and afterwards declines greatly although there are more loss terms.
 Meanwhile, the BLEU scores on the training set keep rising after $\mathcal{L}_\mathrm{c}$ is added till converged. This shows lower training losses correspond to better translation which can be a proof that the training loss is reasonable. 

Our method converged on the 13th epoch and Transformer converged on the 9th epoch. When converged, our method has higher BLEU scores on both the training and valid sets, then We can think our method reaches better optimization than Transformer.

\subsection{The Fluency and Faithfulness of the Translation}

As the evaluation module is designed to consider the fluency and faithfulness of the translation, we assess whether this brings the improvement on the two metrics. For fluency, we test the $n$-gram accuracy on all the test sets where the $n$-gram accuracy  is the ratio of the number of matched $n$-gram between translation and reference against the total number of $n$-gram in the translation. For faithfulness, the cosine similarity between translation and reference is calculated using the average embeddings of all words.

According to the results in Table \ref{tab-ngram}, our method can generate translation with higher $n$-gram accuracy for order $1$ to order $4$ and the difference becomes wider as the order increases. Higher $1$-gram accuracy indicates that our method can reach more ground truth in optimization and this is another proof that the loss we use can lead to better optima. Besides, the greater accuracy on $n$-gram, especially on $3$-gram and $4$-gram, implies better fluency. Furthermore, our method can  have a bigger cosine similarity to the reference and this means the generate translation is more faithful in meaning to the source sentence. In conclusion, our method can produce translation with better fluency and faithfulness.

\section{Conclusions}

\begin{table}[!t]
\centering
\resizebox{1.0\columnwidth}!{
\begin{tabular}{l | c | c | c | c| c}
\hline
& \bf {1-gram} & \bf{2-gram} & \bf{3-gram} & \bf{4-gram} & \bf {Cosine} \\
\hline
\hline
 \bf {Transformer} & 79.10 & 52.72 & 35.34 & 23.82 & 0.873 \\
 \hline
 \bf {Our method} & 79.82 & 54.18 & 36.90 & 25.28 & 0.877 \\
\hline
\end{tabular} 
}
\caption{N-gram accuracy and cosine similarity on CN$\rightarrow$EN translation. N-gram accuracy is the average on all the test sets. Cosine similarity is calculated with the average embeddings of the translation against that of the reference. } \label{tab-ngram}
\end{table}

Teacher forcing employs a cross-entropy loss to supervise the training with a 0-1 distribution and back-propagates gradients only through the ground truth words . When the ground truth word cannot dominate the probability distribution, a major portion of the distribution is discarded, leading to poor optimization. To solve this problem, we introduce an evaluation module to draw a new distribution over all the words and further use the new distribution to guide the training. To make a proper evaluation, we estimate the translation from the perspectives of fluency and faithfulness to appreciate the translation which is fluent in the target and faithful in meaning to the source. The experiments prove that our method can get better performance on multiple data sets with better optimization and meanwhile the generated translation is more fluent in the target and more faithfulness to the source.

\section{Acknowledgements}

We thank all the anonymous reviewers for their insightful and valuable comments.
This work was supported by National Natural Science Foundation of China (NO61876174) and National Key R\&D Program of China (NO2017YFE9132900).


\bibliography{aaai20}
\bibliographystyle{aaai}

\end{document}